\documentclass{article}

\usepackage[nonatbib,final]{nips_2017}

\usepackage[utf8]{inputenc} % allow utf-8 input
\usepackage[T1]{fontenc}    % use 8-bit T1 fonts
\usepackage{booktabs}       % professional-quality tables
\usepackage{amsfonts}       % blackboard math symbols
\usepackage{nicefrac}       % compact symbols for 1/2, etc.
\usepackage{microtype}      % microtypography
\usepackage{color} % DH add

\usepackage{makeidx}  % allows for indexgeneration
\usepackage{amsmath,graphicx,dsfont,bm}
\usepackage{textcomp}

\graphicspath{{.} {./fig/}}

\usepackage{xcolor}
\definecolor{ballblue}{rgb}{0.13, 0.67, 0.8}
\definecolor{dollarbill}{rgb}{0.52, 0.73, 0.4}
\definecolor{indigo}{rgb}{0.0, 0.25, 0.42}

\usepackage[colorlinks=True]{hyperref}

\hypersetup{colorlinks,breaklinks,
            urlcolor=indigo,
            citecolor=indigo,
            linkcolor=indigo}

\newcommand{\abbr}[2]{#1 (#2)}

\newcommand{\norm}{f}
\newcommand{\normaldist}{\mathcal{N}}

\newcommand{\feature}{\mathcal{F}}
\newcommand{\transformer}{\mathcal{T}}
\newcommand{\addgate}{\beta}
\newcommand{\mulgate}{\gamma}

\newcommand{\std}{\text{std}}

\newcommand{\lxl}{\ensuremath{1 \times 1} }

\newcommand{\cX}{\mathcal{X}}
\newcommand{\tcX}{\mathcal{\tilde{X}}}

\newcommand{\zz}{\mathbf{z}}
\newcommand{\yy}{\mathbf{y}}
\newcommand{\xx}{\mathbf{x}}
\newcommand{\txx}{\mathbf{\tilde{x}}}
\newcommand{\R}{\mathds{R}}

\newcommand*\samethanks[1][\value{footnote}]{\footnotemark[#1]}
\newcommand{\inst}[1]{${}^{#1}$}
\begin{document}
\title{Context-based Normalization of Histological Stains using Deep Convolutional Features}
\author{
  D.~Bug\inst{1}\thanks{These authors contributed equally to this work.}\ ,
  S.~Schneider\inst{1}\samethanks\ ,
  A.~Grote\inst{2},
  E.~Oswald\inst{3},
  F.~Feuerhake\inst{2},
  J.~Sch\"uler\inst{3}, 
  D.~Merhof\inst{1} \\
  \inst{1} RWTH Aachen University, Institute for Imaging and Computer Vision, Germany\\
  \inst{2} Hannover Medical School, Institute for Pathology, Germany\\
  \inst{3} Oncotest GmbH, Germany\\
}
% index{Bug, Daniel}
% index{Schneider, Steffen}
% index{Grote, Anne}
% index{Oswald, Eva}
% index{Feuerhake, Friedrich}
% index{Sch\"uler, Julia}
% index{Merhof, Dorit}


\maketitle              % typeset the title of the contribution
\begin{abstract}
While human observers are able to cope with variations in color and appearance of histological stains, digital pathology algorithms commonly require a well-normalized setting to achieve peak performance, especially when a limited amount of labeled data is available.
This work provides a fully automated, end-to-end learn\-ing-based setup for normalizing histological stains, which considers the texture context of the tissue.
We introduce \emph{Feature Aware Normalization}, which extends the framework of batch normalization in combination with gating elements from Long Short-Term Memory units for normalization among different spatial regions of interest.
By incorporating a pretrained deep neural network as a feature extractor steering a pixelwise processing pipeline, we achieve excellent normalization results and ensure a consistent representation of color and texture.
The evaluation comprises a comparison of color histogram deviations, structural similarity and measures the
color volume obtained by the different methods. % change color space size -> color volume
\end{abstract}

%%%
%%% Introduction
%%%

\section{Introduction}
\label{sec:intro}

In digital pathology, histological tissue samples undergo fixation and embedding, sectioning and staining, and are finally digitized via whole-slide scanners.
Each individual processing step presents a potential source of variance. 
In addition to such inherent variation, different staining protocols typically exist between different institutions. 
While trained human observers can handle most of the resulting variability, algorithms typically require a unified representation of the data to run reliably.
Essentially, the challenge in any normalization task is to transform the distribution of color values from an image acquired using an arbitrary staining protocol into a defined reference space, modeled by a different distribution.
Chronologically, normalization algorithms applied in histopathology advanced from color matrix projections \cite{reinhard2001color} to deconvolutional approaches \cite{macenko2009method,vahadane2016structure}, while many modern methods incorporate contextual information \cite{bejnordi2016stain,Janowczyk2017} to improve the normalization result.
This trend is motivated by the amount of internal structures found in histological images, such as cell nuclei or blood-vessels, which may have similar color ranges under varying conditions and thus need to be normalized based on their context rather than individual pixel intensities.
Particularly, the method ``StaNoSa'' \cite{Janowczyk2017}, uses an autoencoder to provide features that steer a histogram-matching, which introduces the tissue context.
In this work, we advance the idea by incorporating the entire normalization process into a deep neural network, where all algorithmic parameters are optimized via stochastic gradient descent. 
This requires several changes to the normalization setting, which we discuss in Sec.~\ref{sec:method}.
We evaluate the approach using an extensive dataset dedicated to the normalization problem, which is described in Sec.~\ref{sec:evaluation}.
The dataset and an implementation of our approach are available at \href{https://github.com/stes/fan}{\texttt{github.com/stes/fan}}.

%%%
%%% Method
%%%

\section{Method}
\label{sec:method}

\begin{figure}[t]%
	\centering
	\includegraphics[width=\textwidth]{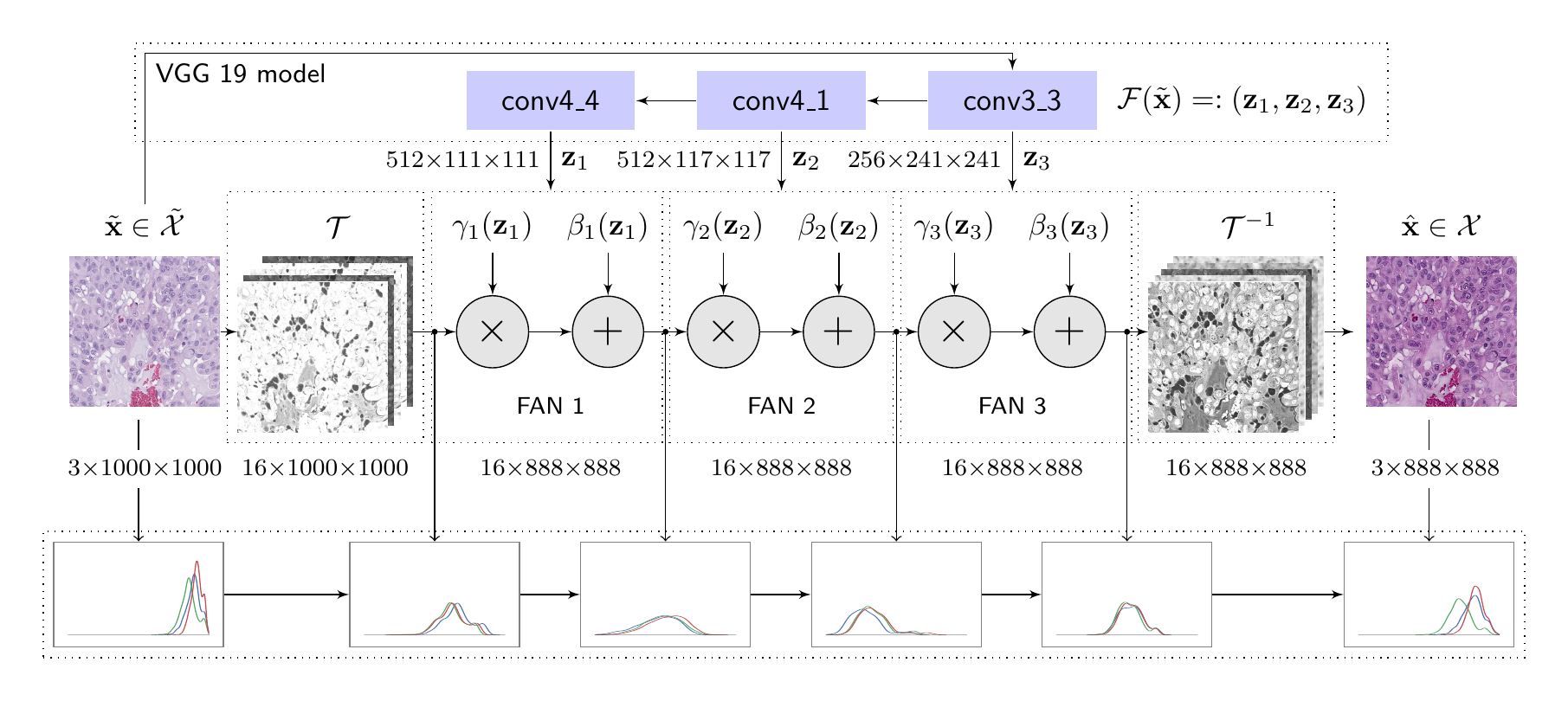}
	\caption{%
Network architecture using Feature-Aware Normalization (FAN):
Multi-scale deep network (here VGG-19) features are computed in a parallel path to the image processing.
In the FAN units, the feature representations from different scales are upsampled to image size and perform pixelwise scaling and shifting of the feature maps computed by $\transformer$ in the main path. Below the network, changes in the RGB histograms are visualized along the path.
Only the valid part of the convolution is used in the feature extractor, which is why the image gets cropped.}%CHANGED: Some Clarification
\label{fig:architecture}%
\end{figure}

The goal of stain normalization is to adapt the color of each pixel based on the tissue context according to visually relevant features in the image.
Similar requirements can be found in artistic style transfer \cite{Dumoulin2016,Gatys2016}, where the characteristics of a particular artistic style, e.g.~van Gogh, are mapped onto an arbitrary image.
However, while artistic style transfer may alter spatial relations between pixels, such as edges and corners, normalization in digital pathology requires changes that are limited to colors while preserving structures.
This is similar to the problem of domain adaptation, where images from arbitrary source domains should be transferred to the target domain  existing algorithms operate on. % CHANGED: added domain transfer view, changed pixel -> spatial pixel

Following this view, we consider manifolds $\cX_i$ representing images from different color protocols.
Furthermore, we assume that each image was generated from a shared underlying latent representation $\Omega$, the image ``content''.
In the process of image acquisition, we consider (unknown) functions $g_i: \Omega \to \cX_i$ that map points from the latent representation to $\cX_i$.
In color normalization, we choose protocol $k$ as a reference $\cX := \cX_k$, while all other datasets are considered to be part of a noise dataset $\tcX := \bigcup_{i \neq k} \cX_i$.
We propose a novel unsupervised learning algorithm capable of computing $f_\theta: \tcX \to \cX$, only using samples from a fixed reference protocol $k$, such that for any latent feature representation $\zz \in \Omega$, $f(\xx_i) = \xx_k$ given that $\xx_i = g_i(\zz)$ and $\xx_k = g_k(\zz)$. % CHANGED: added that only samples from the reference protocol are needed during training

\subsection{Feature-Aware Normalization}

From style transfer, we adapt the idea to use two parallel network paths, as depicted in Fig.~\ref{fig:architecture}.
The main path is a transformer network $\transformer$ along with its inverse $\transformer^{-1}$ performing \lxl convolutions to augment the color space by learning a meaningful latent representation, while a feature extractor $\feature$ provides context information on the parallel path.

As a coupling mechanism we introduce \abbr{Feature-Aware Normalization}{FAN} units, which are inspired by \abbr{Batch Normalization}{BN} \cite{Ioffe2015batch} and the gating mechanism in \abbr{Long Short-Term Memory}{LSTM} units \cite{Hochreiter1997}.
As investigated by \cite{Dumoulin2016}, different styles can be integrated into the same network by adapting the shifting and scaling parameters $\beta$ and $\gamma$ of the BN layers only.
We treat color protocols as a particular simple form of style, restricted to only pixel-wise transformations based on the feature representation $\zz$, which allows changes according to the pixel context, e.g.~nucleus or plasma, while largely preserving the image structure. It is to note that minor structural changes may still occur since despite the restriction, each normalized pixel links back to a perceptive field via $\feature$. % CHANGED: added hints about structural changes

\begin{figure}[t]\centering%
	\includegraphics[width=\textwidth]{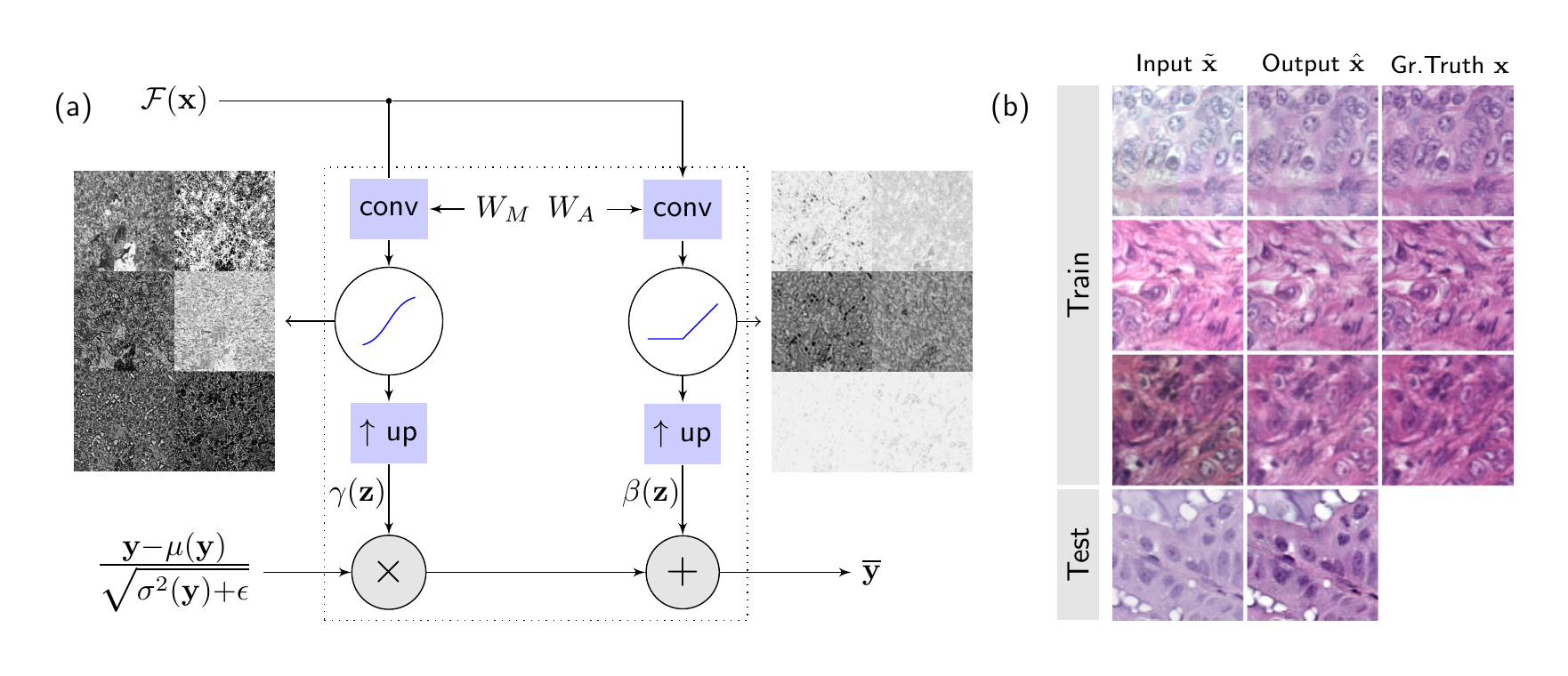}%
	\caption{%
		a) Block diagram of the FAN unit with examples of the pixel wise $\beta$ and $\gamma$ activations next to the nonlinearities.
		b) Three examples of our PCA-based training augmentation and one application case.
	}%
	\label{fig:module}%
\end{figure}

After transforming an input image $\txx$ into the feature spaces $\yy = \transformer(\txx)$ and $\zz = \feature(\txx)$, the normalization procedure for feature map $k$ is given as
\begin{equation}
\overline{\yy}^{(k)} = \frac{\yy^{(k)} - \mu_k(\yy)}{\sqrt{\sigma_k^2(\yy) + \epsilon}}
\cdot \underbrace{\mathrm{sigm}(\sum_j W_M^{(k,j)} \cdot \zz^{(j)})}_{=: \mulgate^{(k)} (\zz)} + 
\underbrace{\max(\sum_j W_A^{(k,j)} \cdot \zz^{(j)}, 0)}_{=: \addgate^{(k)}(\zz)},
\end{equation}
where $\mu_k(\yy)$ and $\sigma_k^2(\yy)$ denote mean and variance computed over the spatial and batch dimension of feature map $k$.
Think of this as a way of error correction, where $\mathrm{sigm} \in (0, 1)$ suppresses incorrect color values and the additive ReLU activation contributes the correction.
A small constant value $\epsilon$ provides numerical stability.
The weight matrices for the multiplication gate $W_M$ and addition gate $W_A$ are used to map the representation from $\feature$ onto the number of feature dimensions provided by $\transformer$.
The result is upsampled to better match the input size, however, some pixel information at the borders is lost due to the use of valid convolutions. % CHANGED: hint about the cropping
A representation of the whole module along with example outputs of $\beta$ and $\gamma$ is depicted in Fig.~\ref{fig:module}. Together with Fig.~\ref{fig:architecture}, this illustrates the basic idea of gradually rescaling and shifting the internal representation.

We use a pretrained VGG-19 architecture with weights from the ILSVRC Challenge \cite{simonyan2014very} as feature extractor $\feature$.
This network was chosen because of its success in style transfer and a variety of classification tasks, which indicates relevant internal features, computed on a dataset whose color variations exceed histopathological stains by far. % CHANGED: hint on histological color variation
However, we stress the fact that competing high scoring architectures, possibly finetuned to the image domain, could easily be deployed in this model as well.
To incorporate context information from various scales, we deploy three FAN units, which is an empirical number.
For $\feature$ to provide meaningful information to the FAN units, we imply that there exists a transformation $\zz' = W \cdot \zz$ such that the result is independent of the staining protocol, i.e., the same tissue stained with two different protocols would yield identical representations in latent variable space, which is a feasible assumption.
Throughout this work, the weights of $\feature$ are fixed and the training process only adapts the parameters of all FAN units and the transformer network $\transformer$.

\subsection{Normalization by Denoising} % CHANGED: Title

Training the network $f_\theta: \tcX \to \cX$ is challenging since no samples $(\xx,\tilde{\xx})$ with matching latent representation $\zz$ are available directly. % CHANGED: slight change of wording: "matching"
To circumvent this issue, we propose the use of a noise model in which the reference image $\xx$ is disturbed by a noise distribution $p(\tilde{\xx} | \xx)$.
During training, for each training example $\xx_i \in \cX$, we use the noise distribution to draw a corresponding sample $\tilde{\xx}_i \sim p(\tilde{\xx} | \xx_i)$, $\tilde{\xx}_i \in \tcX$ and minimize the mean-squared error between the normalized sample and the original one, yielding the objective

\begin{equation}
	\min_\theta \sum_{i = 0}^{N} \| \norm_\theta(\tilde{\xx}_i) - \xx_i \|^2, \quad \tilde{\xx}_i \sim p(\tilde{\xx} | \xx_i) \ \forall i \in [N].
\end{equation}
A minimizer of this objective maps all disturbed images $\tilde{\xx}$ from $\tcX$ onto $\cX$ under the chosen noise model.
For this training scheme to work, $p(\tilde{\xx} | \xx)$ has to assign high probability to samples on $\tcX$.
To ensure this property, we follow the commonly used principle of data augmentation according to the principal components of pixel values.
The noise model is then given by a normal distribution $\normaldist(\xx,  W\Sigma W^T\varepsilon)$.
Herein, $W, \Sigma \in \R^{3\times 3}$ denote the transformation from RGB color values to principal components and the component's explained variance, respectively.
The noise magnitude is controlled by $\varepsilon \in (0,1)$, determined empirically by visual inspection of the samples.
An example of this training augmentation is given in Fig.~\ref{fig:module}b.

%%%
%%% Evaluation
%%%

\section{Experiments}
\label{sec:evaluation}

In order to comprehensively evaluate the normalization performance, we introduce a dedicated dataset comprising five blocks of lung cancer tissue from patient-derived xenografts each of which was sliced consecutively into nine slides.
In addition to variations of the Hematoxylin (H) and Eosin (E) component introduced in \cite{Janowczyk2017}, we vary the slice thickness (T). 
For each slice, the parameters were iterated yielding the staining protocols 1. HET$\uparrow$, 2. HET$\downarrow$, 3. standard HET protocol, 4. HE$\uparrow\,$T, 5. HE$\downarrow\,$T, 6. H$\uparrow\,$ET, 7. H$\downarrow\,$ET, 8. H$\uparrow\,$E$\uparrow\,$T, 9. H$\downarrow\,$E$\downarrow\,$T, where '$\uparrow$' denotes a doubled and '$\downarrow$' a halved concentration or thickness, with the standard values being 1:10 for H (liquid component), 0.6g/200ml for E (powder) and 3\textmu m for T. %CHANGED: slight change of wording --> "slice"
From each digitized whole-slide image (Aperio AT2 Scanner, Leica Biosystems, Wetzlar, Germany), we extracted and manually registered five regions with 1000x1000 pixels, resulting in a test set of 225 images,  wherein the tissue distribution is largely shared between the protocol sets of nine images due to consecutive slicing and registration. % CHANGED: punctuation
For each protocol, a separate normalization network was trained with the particular protocol chosen as the reference dataset $\cX$.
Approximately 15000 patches of size 192$\times$192 were extracted separately from the slides and used for training.
The validation was then done on tissue samples from the remaining protocols, excluding the training patches.

We define three experiments to evaluate the performance of our proposed algorithm in comparison to the previously presented methods of Reinhard (REI, \cite{reinhard2001color}), Macenko (MAR, \cite{macenko2009method}), Vahadane (VAH, \cite{vahadane2016structure}) and Bejnordi (BEJ, \cite{bejnordi2016stain}). For all methods, we repeat the experiments with each of the nine protocols as normalization target, with the exception of BEJ, where we use the provided reference protocol. %CHANGED: removed fill words
Wherever possible, we show the metric computed on the unnormalized images as baseline (BAS).

\begin{figure}[t]\centering%
	\includegraphics[width=\textwidth]{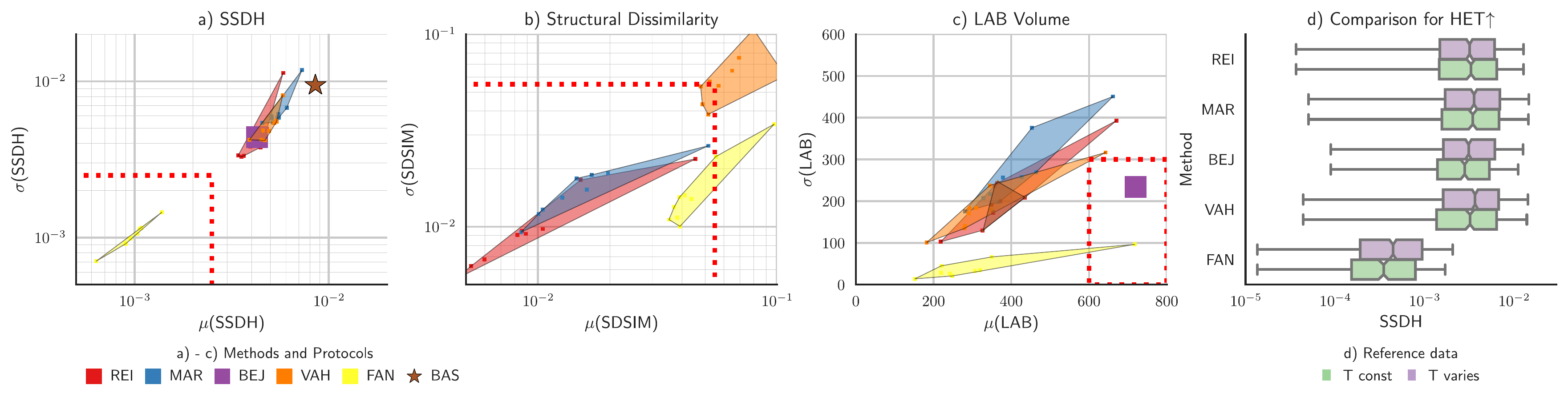}
	\caption{%
Comparison of the methods from Reinhard (REI, \cite{reinhard2001color}), Macenko (MAR, \cite{macenko2009method}), Vahadane (VAH, \cite{vahadane2016structure}), Bejnordi (BEJ, \cite{bejnordi2016stain}) with the proposed FAN method. The metrics for unnormalized images are given as baseline (BAS) in case of SSDH and LAB. Plots a)-c) show the convex hulls of the metric distributions for each algorithm. Red dotted lines highlight preferred regions (best viewed in the digital version).}%
	\label{fig:metrics}%
\end{figure}

\textbf{1. Elimination of protocol deviations.}
As suggested by related publications, we measure distances between color distributions of registered patches in consecutive slices of tissue. 
As an estimator for the true distribution, we compute a kernel-density estimate (KDE) on a 256-bin histogram of each color channel.
A binomial filter of length seven is used as KDE kernel.
For clarity, we now slightly deviate from previous notation and regard $\xx_i$ as a patch from protocol $i$ and $f_j$ as the normalization function to reference dataset $j$.
This yields the metric $\ell_{i, j} = \mathrm{SSDH}(f_j(\xx_i), \xx_j)$, where SSDH denotes the sum of squared differences between the color histograms of $f_j(\xx_i)$ and $\xx_j$.
In Fig.~\ref{fig:metrics}a the error distribution is visualized as a convex hull plot, where each point inside the hull represents the performance of a normalization $f_j$ which is characterized by the mean and standard-deviation of all corresponding $\ell_{i, j}$. % CHANGED: clarify again what we are plotting

\textbf{2. Influence on texture.}
Whereas pixelwise methods do usually not interfere with pixel relations, introducing context into the normalization carries the risk of erroneous alterations in groups of pixels, since output pixels connect to a perceptive field via $\feature$.
Hence, we apply the structural dissimilarity index (SDSIM, \cite{wang2004image}) as a metric, which quantifies perceptual dissimilarity by comparing first- and second-order statistics of the images before and after normalization. 
A value of zero indicates identical image structure, while higher values indicate dissimilarity.
This yields the metric $\ell_{i, j} = \mathrm{SDSIM}(f_j(\xx_i), \xx_i)$ shown in Fig.~\ref{fig:metrics}b. % CHANGED: replaced some words to adjust length of the paragraph

\textbf{3. Color richness.}
Theoretically, a normalization effect with respect to a histogram deviation metric can be erroneously created by computing a lowly saturated subspace with correlated channels, which basically leads to a grayscale-like appearance.
To reassure a valid normalization, we use the volume of colors in Lab color space as a measure for the amount of perceivable colors after normalization.
Since all components of the Lab space are decorrelated, the approximate color volume is computed as product of the standard deviations in each channel  $\mathrm{LAB} := \std(L)\cdot\std(a)\cdot\std(b)$.
This yields the metric $\ell_{i, j} = \mathrm{LAB}(f_j(\xx_i))$, shown in Fig.~\ref{fig:metrics}c.
Ideally, a high color volume is achieved at a low variance. % CHANGED: slight change of wording to adapt the length of the paragraph

\begin{figure}[t]\centering%
\includegraphics[width=0.9\textwidth]{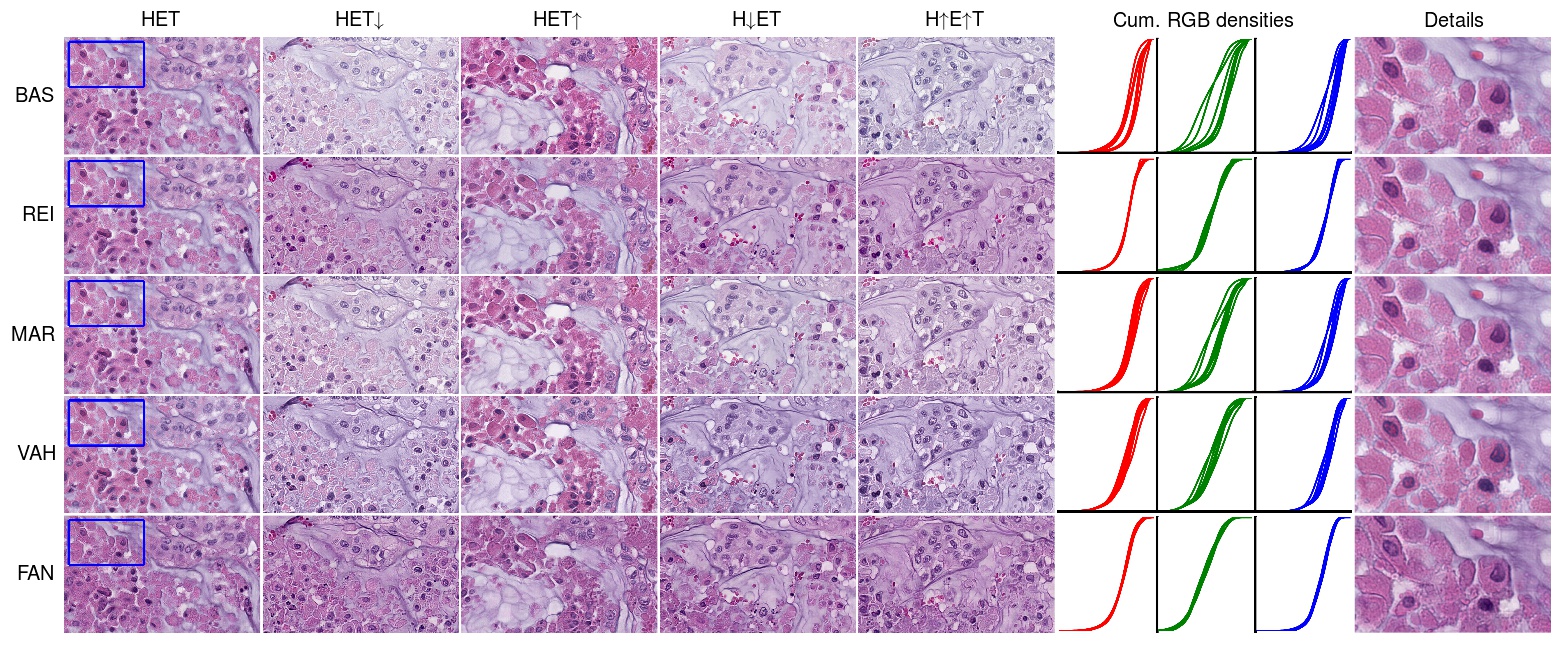}
\caption{%
Normalization of selected source protocols together with cumulated density per channel (all protocols) and detail view. Note that each algorithm is shown with the best performing target protocol according to SSDH and LAB.
}
\label{fig:images}%
\end{figure}

%%%
%%% Discussion
%%%

\section{Discussion}
\label{sec:discussion}

From quality metrics in Fig.~\ref{fig:metrics} and the overview in Fig.~\ref{fig:images}, we conclude that in terms of SSDH and color volume our proposed method outperforms previous approaches by a notable margin. % CHANGED: color volume is also better -> lower variance
As expected, the introduction of context information may result in minor texture changes, which originate from the different scales of the FAN units and the choice of $\feature$ and display as faint ``tiles'' in the output image.
According to visual inspection, e.g.~in the detailed views in Fig.~\ref{fig:images}, these effects can be considered negligible.
In many cases, the volume of perceivable colors decreases in FAN compared to conventional methods.
However, the low variance in the color volumes indicates a strong consistency of the resulting color spaces.
In contrast, the high variance in some cases of the MAR method results from artifacts where a failed estimation of source or target color space leads to unrealistically intense and oversaturated colors.
We may also note that the performance of both MAR and REI strongly depends on having similar tissue distributions in the source and reference patches, which is a property of the used dataset, but not given in general. % CHANGED: slight change of wording
Hence, the performance we report for these algorithms can be seen as an upper bound.
VAH achieves a very reliable color normalization, but structures can be affected if the sparse decomposition does not find a good optimum.
With the HET$\uparrow$ protocol as target, we obtain a model which shows an exceptionally large color volume.
In application, not every target protocol is equally important. 
% Since normalization is often applied to cope with unsaturated colors and to increase contrast for subsequent processing steps, one exceptionally good color normalizer is a very satisfying outcome with a high practical value. % CHANGED: Limited relevance? leave out?

With the proposed approach, we overcome the challenge of providing a suitable reference image covering a variety of tissue statistics by finding an appropriate target protocol with high color volume and contrast. These properties are then encoded in the network parameters through training.
As the low variance in all measures shows, the normalization characteristics encoded in FAN units exceed the capabilities of previous methods.
We find further indication for this by comparing moderate stain variation (protocols 3-9) with deviations in thickness (protocols 1-3) in Fig.~\ref{fig:metrics}d. 
Herein, previous methods show a notable decay in the quality metrics, whereas FAN provides a very consistent normalization.

A beneficial advantage of the clear separation between the feature extractor and the transformer path is that the currently used VGG-19 architecture can easily be replaced by any other feature extractor.
Particularly, computationally more efficient network structures with less parameters are good candidates. % CHANGED: add a hint that finetuned nets can be easily incorporated
Additionally, it might be promising to train the feature extraction step jointly with the remaining network, which is left for further research.

In a greater perspective, the general approach to separate network functions into distinct parts may have application to other areas as well, such as the artistic style transfer networks that originally motivated our work.
We assume that an artistic style transfer can be realized with this architecture simply by using larger filter sizes in the transformer network $\transformer$, allowing for spatial context aggregation on this path.

%%%
%%% References
%%%

\small
\bibliographystyle{splncs03}
\bibliography{ref}

\begin{thebibliography}{10}
\providecommand{\url}[1]{\texttt{#1}}
\providecommand{\urlprefix}{URL }

\bibitem{bejnordi2016stain}
Bejnordi, B.E., Litjens, G., Timofeeva, N., Otte-H{\"o}ller, I., Homeyer, A.,
  Karssemeijer, N., van~der Laak, J.A.: {Stain Specific Standardization of
  Whole-Slide Histopathological Images}. IEEE Transactions on Medical Imaging
  35(2),  404--415 (2016)

\bibitem{Dumoulin2016}
Dumoulin, V., Shlens, J., Kudlur, M.: A learned representation for artistic
  style. CoRR  abs/1610.07629 (2016)

\bibitem{Gatys2016}
Gatys, L.A., Ecker, A.S., Bethge, M.: {Image Style Transfer Using Convolutional
  Neural Networks}. In: The IEEE Conference on Computer Vision and Pattern
  Recognition (CVPR) (2016)

\bibitem{Hochreiter1997}
Hochreiter, S., Schmidhuber, J.: {Long Short-Term Memory}. Neural Computation
  9(8),  1735--1780 (1997)

\bibitem{Ioffe2015batch}
Ioffe, S., Szegedy, C.: {Batch Normalization: Accelerating Deep Network
  Training by Reducing Internal Covariate Shift}. In: Proceedings of the 32nd
  International Conference on Machine Learning (ICML-15). pp. 448--456 (2015)

\bibitem{Janowczyk2017}
Janowczyk, A., Basavanhally, A., Madabhushi, A.: {Stain Normalization using
  Sparse AutoEncoders (StaNoSA): Application to Digital Pathology}.
  Computerized Medical Imaging and Graphics  (Feb 2017)

\bibitem{macenko2009method}
Macenko, M., Niethammer, M., Marron, J.S., Borland, D., Woosley, J.T., Guan,
  X., Schmitt, C., Thomas, N.E.: {A method for normalizing histology slides for
  quantitative analysis}. In: 2009 IEEE International Symposium on Biomedical
  Imaging: From Nano to Macro. pp. 1107--1110 (June 2009)

\bibitem{reinhard2001color}
Reinhard, E., Adhikhmin, M., Gooch, B., Shirley, P.: {Color Transfer between
  Images}. IEEE Computer Graphics and Applications  21(5),  34--41 (2001)

\bibitem{simonyan2014very}
Simonyan, K., Zisserman, A.: {Very Deep Convolutional Networks for Large-Scale
  Image Recognition}. CoRR  abs/1409.1556 (2014)

\bibitem{vahadane2016structure}
Vahadane, A., Peng, T., Sethi, A., Albarqouni, S., Wang, L., Baust, M.,
  Steiger, K., Schlitter, A.M., Esposito, I., Navab, N.: {Structure-Preserving
  Color Normalization and Sparse Stain Separation for Histological Images}.
  IEEE Transactions on Medical Imaging  35(8),  1962--1971 (2016)

\bibitem{wang2004image}
Wang, Z., Bovik, A.C., Sheikh, H.R., Simoncelli, E.P.: {Image Quality
  Assessment: From Error Visibility to Structural Similarity}. IEEE
  Transactions on Image Processing  13(4),  600--612 (2004)

\end{thebibliography}

\end{document}